%% file: aaai23.tex
\documentclass[letterpaper]{article} 
\usepackage{aaai23}  
\usepackage{times}  
\usepackage{helvet}  
\usepackage{courier}  
\usepackage[hyphens]{url}  
\usepackage{graphicx} 
\usepackage{natbib}  
\usepackage{caption} 
\usepackage{algorithm}
\usepackage{algorithmic}

\usepackage{url}            
\usepackage{booktabs}       
\usepackage{amsfonts}       
\usepackage{nicefrac}       
\usepackage{microtype}      
\usepackage{xcolor}         
\usepackage{graphicx}
\usepackage{amsmath}
\usepackage{amssymb}
\usepackage{amsfonts}
\usepackage{multirow}
\usepackage{color}
\usepackage{colortbl}
\usepackage{wrapfig}
\definecolor{mygray}{gray}{.92}
\definecolor{myblue}{RGB}{240,248,255}
\definecolor{myred}{RGB}{255,228,225}

\DeclareMathOperator{\softmax}{Softmax} 

\DeclareMathOperator{\MLP}{MLP}
 
\DeclareMathOperator{\Sim}{Sim}

\DeclareMathOperator{\f}{\mathbf{f}}
\DeclareMathOperator{\fp}{\mathbf{f'}} 
\DeclareMathOperator{\et}{\epsilon} 

\title{Self-Supervised Video Representation Learning via Latent Time Navigation}

\def\ie{\emph{i.e.}}
\def\eg{\emph{e.g.}}

\def\wrt{\emph{w.r.t.}}

\usepackage{newfloat}
\usepackage{listings}
\DeclareCaptionStyle{ruled}{labelfont=normalfont,labelsep=colon,strut=off} 
\floatstyle{ruled}
\newfloat{listing}{tb}{lst}{}
\floatname{listing}{Listing}
\author{
    Di Yang\textsuperscript{\rm 1,2}, 
    Yaohui Wang\textsuperscript{\rm 1,2,5\thanks{Work done while the author was at Inria}}, 
    Quan Kong\textsuperscript{\rm 4}, 
    Antitza Dantcheva\textsuperscript{\rm 1,2}, 
    Lorenzo Garattoni\textsuperscript{\rm 3}, \\
    Gianpiero Francesca\textsuperscript{\rm 3}, 
    François Brémond\textsuperscript{\rm 1,2} 
}
\affiliations{
    \textsuperscript{\rm 1}Inria, 2004 Rte des Lucioles, {Valbonne}, France \\ 
    \textsuperscript{\rm 2}Université Côte d'Azur, 28 Av. de Valrose, {Nice}, {France} \\
    \textsuperscript{\rm 3}Toyota Motor Europe, 60 Av. du Bourget, {Brussels}, {Belgium} \\ 
    \textsuperscript{\rm 4}Woven Planet Holdings, 3-2-1 Nihonbashimuromachi, Chuo-ku, Tokyo, Japan \\ 
    \textsuperscript{\rm 5}Shanghai AI Laboratory, 701 Yunjin Road, Shanghai, {China} \\ 
    
{\{di.yang, yaohui.wang, antitza.dantcheva, francois.bremond\}@inria.fr}, 

{\{lorenzo.garattoni, gianpiero.francesca\}@toyota-europe.com, 
quan.kong@woven-planet.global}
}

\usepackage{bibentry}

\begin{document}

\maketitle

\begin{abstract}

Self-supervised video representation learning aimed at maximizing similarity between different temporal segments of one video, in order to enforce feature persistence over time. This leads to loss of pertinent information related to temporal relationships, rendering actions such as `enter' and `leave' to be indistinguishable.
To mitigate this limitation, we propose Latent Time Navigation (LTN), a time-parameterized contrastive learning strategy that is streamlined to capture fine-grained motions. Specifically, we maximize the representation similarity between different video segments from one video, while maintaining their representations \textit{time-aware} along 
a subspace of the latent representation code including an orthogonal basis to represent temporal changes.
Our extensive experimental analysis suggests that learning video representations by LTN 
consistently improves performance of action classification in fine-grained and human-oriented tasks (\eg, on Toyota Smarthome dataset). In addition, we demonstrate that our proposed model, when pre-trained on Kinetics-400, generalizes well onto the unseen real world video benchmark datasets UCF101 and HMDB51, achieving state-of-the-art performance in action recognition.
\end{abstract}

\input{sections/introductionNEW}

\input{sections/related_work}

\input{sections/method}

\input{sections/experiments}

\input{sections/conclusion}

\bibliography{aaai23}
\end{document}

%% file: sections/introductionNEW.tex
 \section{Introduction}
\input{images/intro}

Contrastive learning~\cite{contrastive} is a prominent variant in learning self-supervised visual representations. 
The associated objective is to minimize the distance between latent representations of positive pairs, while maximizing the distance between latent representations of negative pairs. For instance, a visual encoder aims at learning the invariance of multiple \textit{views} of a scene, which constitute positive pairs, by extracting generic features of images~\cite{bachman2019learning, SwAV, chen2020simclr, byol,  He_2020_CVPR, hjelm2019learning, jiao2020subgraph,  tian2020cmc, non-para} or videos~\cite{study2021, cotraining, contras1, kong2020cyclecontrast, ConNTU, Li_2021_ICCV,  gaussian2022, orvpe, yang2022via, Sun_2021_ICCV}. Then, the trained visual encoder can be transferred onto other downstream tasks. 

Remarkable results have been reported by augmentation-invariant contrastive learning. 
In this context, contrastive learning methods enable the visual encoder to find compact and meaningful image representations, invariant to data augmentation. The latent representation of two augmented views of the same instance are enforced to be similar via contrastive learning. In \textit{image-based tasks}, a common augmentation method relates to random cropping~\cite{chen2020simclr, non-para}. 
When extending this idea to \textit{videos}, which are endowed with additional temporal information, cropping in the spatial dimension~\cite{kong2020cyclecontrast} is not sufficient for training an effective visual encoder. Therefore, recent works~\cite{study2021, Li_2021_ICCV, Sun_2021_ICCV} sample different views with a \textit{temporal shift}, learning representations that are invariant to time changes. 
However, for downstream tasks involving temporal relationships, a representation invariant to temporal shifts might omit valuable information. 
For instance, in differentiating actions such as `enter' and `leave' the temporal order is fundamental. 
Hence, a trained visual encoder remains a challenge in handling downstream video understanding tasks such as fine-grained human action recognition~\cite{Das_2019_ICCV, somethingsomething, uav}.

Motivated by the above, we propose Latent Time Navigation (LTN), a time parameterization scheme streamlined to learn time-aware representations on top of the contrastive module. 
As illustrated in Fig.~\ref{fig:intro}, deviating from current contrastive methods~\cite{study2021, He_2020_CVPR, tian2020cmc, non-para} which directly maximize the similarity between representations obtained from the visual encoder for positive samples, LTN encompasses the following steps. 
Firstly, 
we decompose a subspace (\ie, a learnable orthogonal basis and associated magnitudes) from the latent representation code for the video segment, namely `time-encoded component', to do with temporal changes (\eg, changes in appearances, motion, object locations). The other subspace (`time-invariant component') has to do with invariant information. Subsequently, we embed the \textit{time shift value} used for generating data view into a high-dimensional vector as the magnitudes of the directions in the orthogonal basis and then encode this time information into the `time-encoded component' by linear combination of the orthogonal basis and the magnitudes. Finally, we conduct contrastive learning on the entire time-parameterized representations in order to maximize the similarity between positive pairs along the `time-invariant component', while maintaining their representations \textit{time-aware} along the `time-encoded component'.
We note that LTN incorporates time information for video representations and therefore is able to model subtle motions within an action. Consequently, the time-aware representation obtained from the trained visual encoder generalizes better to unseen action recognition datasets, especially to our target human-oriented fine-grained action classification dataset~\cite{Das_2019_ICCV}.

In summary, the contributions of this paper include the following.
(a) We propose Latent Time Navigation (LTN) to parameterize the time information (used for generating data views) on top of contrastive learning, in order to learn a \textit{time-aware} video representation. 
(b) We demonstrate that LTN can effectively learn the consistent amount of temporal changes with the video segments on the decomposed `time-encoded components'.
(c) We set a new state-of-the-art with LTN on the real world dataset (\eg, Toyota Smarthome) for fine-grained action recognition with self-supervised action representation learning.
(d) We demonstrate that our proposed model, when pre-trained on Kinetics-400 dataset, generalizes well to unseen real-world video benchmarks (\eg, UCF101 and HMDB51) with both linear evaluation and fine-tuning.

%% file: images/intro.tex
\begin{figure*}[t]
\begin{center}
\includegraphics[width=0.88\linewidth]{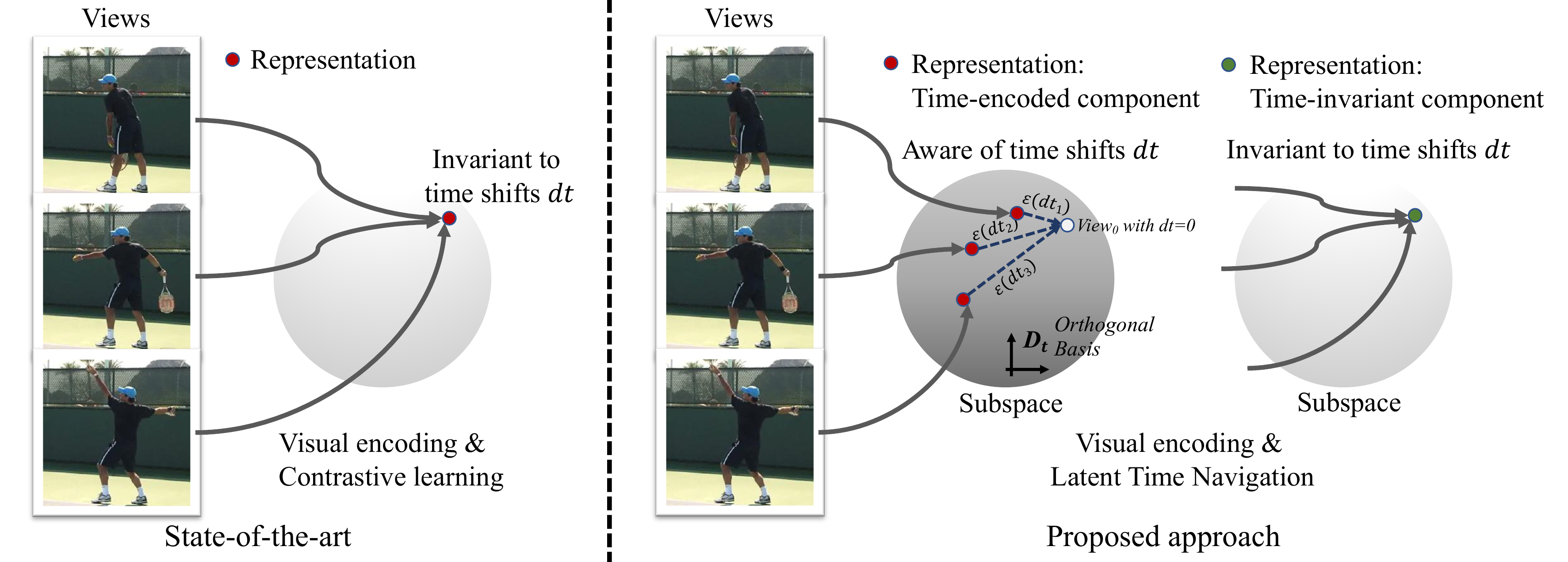}
\end{center}
\vspace{-0.5cm}
   \caption{Current methods (left) leverage on contrastive learning to maximize representation similarities of multiple positive views (segments with time spans and data augmentation) of the same video instance to represent them as a consistent representation. To further improve the representation capability for fine-grained tasks without losing important motion variance, our approach (right) incorporates a time-parameterized contrastive learning (LTN) to remain the video representations aware to time shifts (starting time) in a decomposed time-encoded subspace.}
\vspace{-.3cm}
\label{fig:intro}
\end{figure*}

%% file: sections/related_work.tex
\section{Related Work}
\input{images/overview}

\paragraph{Contrastive Learning.}
Contrastive learning and its variants~\cite{bachman2019learning, SwAV, chen2020simclr, byol, He_2020_CVPR, hjelm2019learning, jiao2020subgraph, tian2020cmc, non-para} have established themselves as a pertinent direction for self-supervised representation learning for a number of tasks due to promising performances. Recent video representation learning methods~\cite{study2021, contras1, kong2020cyclecontrast} are inspired by image techniques. The objective of such techniques is to encourage representational invariances of different views (\ie, positive pairs) of the same instance obtained by data augmentation, \eg, random cropping~\cite{chen2020simclr, non-para}, rotation~\cite{rotation}, while spreading representations of views from different instances (\ie, negative pairs) apart. To further improve the representation capability, CMC~\cite{rotation} scaled contrastive learning to any number of views. MoCo~\cite{He_2020_CVPR} incorporated a dynamic dictionary with a queue and a moving-averaged encoder. To omit a large number of negative pairs, BYOL~\cite{byol} and SwAV~\cite{SwAV} were targeted to solely rely on positive pairs. However, these methods miss a crucial Time element when they are straightforward applied to the \textit{video} domain with views generated by \textit{image} data augmentation technique. In our work, we adopt recent contrastive learning frameworks~\cite{byol,He_2020_CVPR} and we focus on learning time-aware representations for videos by latent spatio-temporal decomposition and navigation in the representation space.

\vspace{-0.1cm}
\paragraph{Self-supervised Video Representation Learning.}
Approaches for self-supervised video representation learning exploit spatio-temporal pretext tasks from numerous unlabeled data. Towards effective extraction of the pertinent motion information in the time dimension, a number of temporal pretext tasks were proposed, \eg, pixel-level future generation~\cite{adv2, pixelpred, adv1, pixelpred2} and jigsaw-solving~\cite{jigsaw-solving}. Additionally, in order to facilitate the learning process, numerous works focused on learning representations in a more abstract space including temporal order~\cite{order, cliporder} or arrow~\cite{arrow} prediction of video frames, future prediction~\cite{future}, speed prediction~\cite{speed}, motion prediction~\cite{motion} and a combination of these tasks~\cite{taco}. These methods are highly constrained by the limited quality of pretext tasks. Recently, video contrastive learning methods~\cite{contras1, kong2020cyclecontrast} have obtained promising results and a large-scale study~\cite{study2021} has been conducted to compare state-of-the-art image-based contrastive methods~\cite{SwAV, chen2020simclr, byol, He_2020_CVPR} on videos using spatio-temporal cropping, color jitters and Gaussian blur data augmentation techniques to generate multiple video views. Further, to improve representation performance, \cite{ding2022motion, selfrgb_2021_huang, transformer2022selfsupervised} focused on view generation techniques, \eg, context-motion decoupling~\cite{selfrgb_2021_huang}, foreground-background merging~\cite{ding2022motion}, global and local sampling across space and time~\cite{transformer2022selfsupervised}. In addition, some specific designs are incorporated in spatio-temporal representation learning including Gaussian probabilistic representations~\cite{gaussian2022}, skeleton contrastive learning~\cite{ConNTU, orvpe, das2021vpn+} and muti-modal learning with audio~\cite{audio2, audio1, audio4, broaden_2021_ICCV, audio3, audio5} or with optical flow~\cite{cotraining, Li_2021_ICCV}. Such contrastive methods aimed at learning video representations invariant to time shift. However, motion significantly changes with time shifts, leading to poor performance on downstream fine-grained action recognition tasks that highly rely on the motion variance. To address this issue, CATE~\cite{Sun_2021_ICCV} proposed to parameterize data augmentation relying on an additional Transformer head prior to contrastive learning. It demonstrated that awareness of the temporal data augmentation is particularly instrumental in fine-grained action recognition tasks. Deviating from CATE that shifts the entire visual representation along all dimensions by the time-shift values even for the action with small motion variances, we study variant time-parameterization strategies and propose to encode the time-shift values partially on certain orthogonal directions instead of on the entire visual representation. By our proposed LTN, the impact of time can be video specific and controlled by the number of the orthogonal directions so that the visual encoder can better capture motions.

%% file: images/overview.tex
\begin{figure*}[t]
\begin{center}
\includegraphics[width=0.93\linewidth]{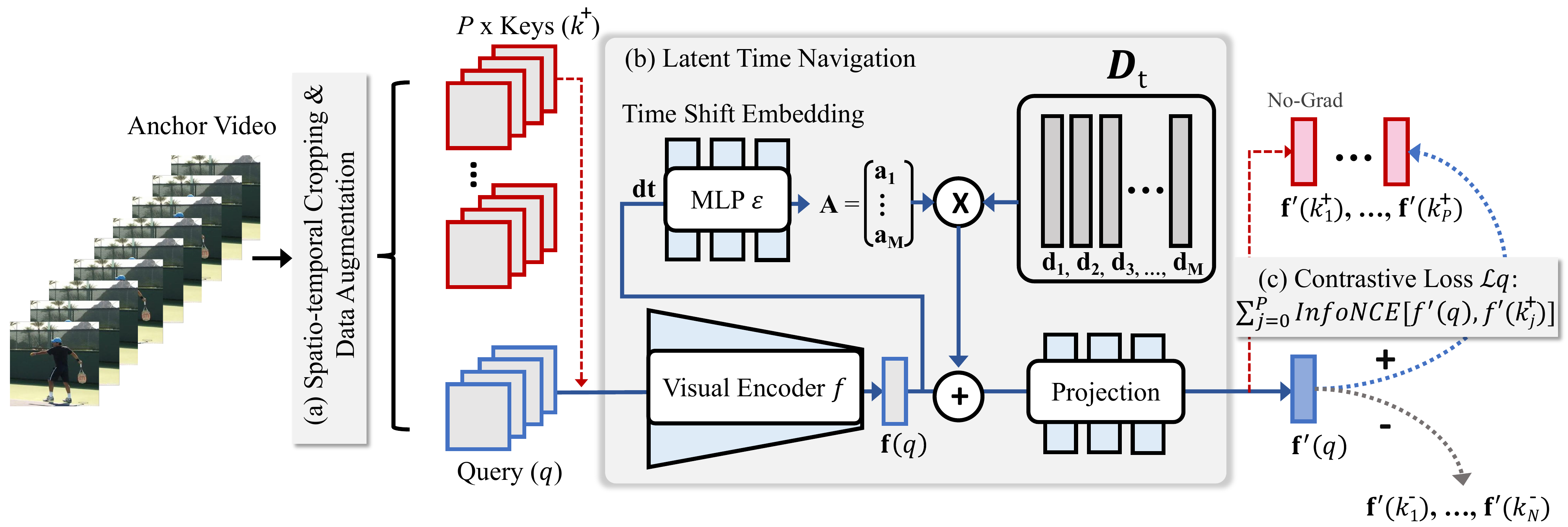}
\end{center}
\vspace{-0.4cm}
   \caption{\text{Overview of the proposed LTN framework.} At each training iteration, given an input video, (a) a query clip ($q$) and multiple positive key clips ($k^{+}_1, k^{+}_2, ..., k^{+}_P$) are generated by data augmentation with different temporal shifts $\mathbf{dt}$. All clips are then fed to a visual encoder that extracts spatio-temporal features for each clip. To learn time-aware representations for query and key clips, (b) we first pre-define a learnable orthogonal basis $\mathbf{D}_t$ ($\mathbf{d_1}, \mathbf{d_2}, ..., \mathbf{d_M}$) that represents the `time-encoded component'. The video representations are expected to be time-aware along $\mathbf{D}_t$ in the training stage. To do so, we transform each query and key video representation (\ie, $\f (q)$, $\f (k^{+}_p)$) by a linear combination of $\mathbf{D}_t$ and associated magnitudes learned from its time shift $\mathbf{dt}$ to a time-blended position (\ie, $\fp (q, \mathbf{dt}_q)$, $\fp (k^{+}_p,  \mathbf{dt}_p)$, abbreviated as $\fp (q)$, $\fp(k^{+}_p)$). Finally, we conduct (c) contrastive learning on top of $\fp$, so that the learned representation from the visual encoder can maintain temporal awareness.}
\vspace{-.25cm}
\label{fig:overview}
\end{figure*}

%% file: sections/method.tex
\section{Proposed Approach}

In this section we introduce our Latent Time Navigation (LTN) framework. We start with the overall architecture, then we proceed to describe the design strategies focusing on time parameterization that enforces the learned video representation to be aware of motion variances.

\subsection{Overall Architecture of LTN}
Our objective is to train a generic visual encoder $\f$ for extracting accurate spatio-temporal features of video clips. We design our visual encoder to be efficient for downstream fine-grained action recognition tasks. We illustrate the overview of the architecture in Fig.~\ref{fig:overview}. To train the visual encoder, a general data augmentation technique including random temporal shifts is applied to generate multiple positive views for a given input video, allowing us to obtain multiple representations from different views. Deviating from previous methods~\cite{He_2020_CVPR, tian2020cmc}, which directly employ contrastive learning 
for these representations in order to make them invariant to spatio-temporal augmentation, we design an additional time parameterization module to blend temporal augmentation to a `time-encoded component' prior to contrastive learning. We then perform the contrastive learning for the new time-blended representations in the training stage. The trained visual encoder can thus be aware of time shifts compared to other positive pairs and can capture the important motion variances of videos for improving fine-grained action recognition tasks.

\vspace{-0.1cm}
\paragraph{View Generation and Embedding.}
Following the study \cite{study2021}, we first spatio-temporally crop a segment by randomly selecting a segment and cropping out a fixed-size box from the same video instance. We then pull together image-based augmentations including random horizontal flip, color distortion and Gaussian blur following~\cite{chen2020simclr, He_2020_CVPR} to generate positive views of the input video at each training iteration. As demonstrated in~\cite{study2021}, multiple positive samples with large time spans between them are beneficial in downstream performance. In our work, we sample a query clip noted as $q$ and multiple positive keys with large time spans, noted as $k^{+}_1, ..., k^{+}_P$ (see Fig.~\ref{fig:overview} (a)). We utilize a 3D-CNN network~\cite{3d-resnet} as the visual encoder to obtain $dim$-dimensional representations of all clips (\ie, $\f(q), \f(k^{+}_1), ..., \f(k^{+}_P) \in \mathbb{R}^{1 \times dim}$). 

\vspace{-0.1cm}
\paragraph{Awareness of Time in Latent Space.}
Large time spans between positive samples may depict significant changes in human motion. When directly matching $\f(q)$ to all positive pairs, the corresponding representations may lose pertinent motion variance caused by time shifts. This could compromise the accuracy of downstream tasks related to fine-grained human motion (\eg, classification of `Leave/Enter', `Stand up/Sit down'). Hence, we expect positive pairs to be partially similar with each other (due to static object, scene) while also partially aware of their time shifts to preserve temporal dynamic information (e.g., changes in motion). To do so, we design several time parameterization methods (see Sec.~Time Parameterization in Latent Space) to encode the time shift value (denoted as $\mathbf{dt_q}$ for $q$) used for data augmentation to a part (several orthogonal directions) of the visual representation while keep the remaining part unchanged. Such time-encoded pretext representation of $q$ and each positive key can be computed and denoted as $\fp (q, \mathbf{dt_q})$ and $\fp (k^{+}_p, \mathbf{dt_p})$. We then maximize the mutual information between the pretext representations $\fp (q, \mathbf{dt_q})$ and $\fp (k^{+}_p, \mathbf{dt_p})$ by contrastive learning.
The original (target) visual representations from different segments (\eg, $\f (q)$, $\f(k^{+}_p)$) will be sensitive to time along the time-encoded part 
after learning and can be transferred onto downstream tasks.

\subsection{Time Parameterization in Latent Space}\label{sec:approaches}
We first introduce the latent space decomposition approach to split the representation space into `time-encoded component' and `time-invariant components', and then we introduce time encoding which is used as a parameter to transform the visual representation only along the `time-encoded component' to reach a new time-blended position.

\vspace{-0.1cm}
\paragraph{Latent Space Decomposition.}\label{sec:time-rep}
To decompose the representation space, we set a learnable orthogonal basis (\ie, a subspace) $\mathbf{D}_t = \{ \mathbf{d_1}, \mathbf{d_2}, ..., \mathbf{d_M} \}$ with $M \in [1, dim)$, and $\mathbf{d} \in \mathbb{R}^{dim \times 1}$ to represent the `time-encoded component', where each vector indicates a basic visual transformation. Due to $\mathbf{D}_t$ entailing an orthogonal basis, any two directions $\mathbf{d_i}, \mathbf{d_j}$ follow the constraint in Eq.~\ref{D_t}.
We implement $\mathbf{D}_t  \in \mathbb{R}^{dim \times M}$ as a learnable matrix following \cite{wang2022latent}, and we apply the Gram-Schmidt algorithm during each forward pass in order to satisfy the orthogonality. 
\begin{equation}\label{D_t}
\small
<\mathbf{d_i}, \mathbf{d_j}>=\left\{\begin{matrix}
 0&i \neq j\\ 
 1&i = j.
\end{matrix}\right.
\end{equation}

\vspace{-0.1cm}
\paragraph{Time Encoding.}
We decomposed the `time-encoded component' $\mathbf{D}_t$ of the video representation from the latent space, to force the model to be aware of temporal variances along $\mathbf{D}_t$ with different time shifts. We propose to encode and parameterize the time shift values $\mathbf{dt}$ for the randomly selected query (and key) segment using their absolute starting point in seconds in the timestamps (\ie, $t_{start}$). We used absolute time as we aimed at learning the representation of a single segment aware of time shift from a fixed `reference view' (\ie, the video beginning).
\begin{equation}\label{dt}
\small
\et (\mathbf{dt}, \f (q)) = \MLP \big( [\MLP(t_{start}), \f (q)]\big).
\end{equation}
Specifically, 
we encode $\mathbf{dt}$ into a high-dimensional vector $\et (\mathbf{dt}, \f (q))$ by simple MLP (see Eq.~\ref{dt}), with the purpose of parameterizing the time shift considering different time-blending variants followed by contrastive learning. 
The time encoder also accepts $\f (q)$ as the input by concatenating with embedded $\mathbf{dt}$, towards learning a video specific encoding. We explore the idea of effective modeling for time shifts by proposing and comparing three time parameterization variants for the transformation from $\f$ to $\fp$.
The first approach has to do with straightforward linear addition on video representation $\f (q)$ with $\et (\mathbf{dt}, \f (q))$ (Variant 1). We then develop more efficient variants, which model the `time-encoded component' more finely by learning the weights (Variant 2) or the magnitudes (Variant 3) only along the directions in the `time-encoded component' $\mathbf{D}_t$. %

\vspace{-0.1cm}
\paragraph{Variant 1. Time-driven Linear Addition}
We implement $\et (\mathbf{dt_q}, \f (q)) \in \mathbb{R}^{1 \times dim}$ as the offset, from which positive pairs need to be pulled away from the representation `time-encoded component' to obtain the time-blended representation in the latent space. The linear addition can be described as Eq.~\ref{linear}.  
\begin{equation}\label{linear}
\small
\fp (q, \mathbf{dt}_q) = \f (q) + \et (\mathbf{dt_q}, \f (q))
\end{equation}

\vspace{-0.1cm}
\paragraph{Variant 2. Time-driven Attention}
We then explicitly implement an attention mechanism to learn a set of attention weights for the positive pairs to be driven by $ \mathbf{W} \in \mathbb{R}^{1 \times M} = \{w_{1}, w_{2}, ..., w_{M}\} = \softmax \big( \et (\mathbf{dt_q}, \f (q)) \big)$. The attention weights force $\f (q)$ to focus on the specific `time-encoded component' in $\mathbf{D}_t$ according to different time encoding. This process can be described as follows
\begin{equation}\label{att}
\small
\fp (q, \mathbf{dt}_q) = \f (q) \cdot (\sum^{M}_{i=1}w_{i} \cdot \mathbf{d_{i}}) .
\end{equation}

\vspace{-0.1cm}
\paragraph{Variant 3. Time-driven Linear Transformation.}
As shown in Fig.~\ref{fig:overview} (b), we finally propose a linear transformation method to encode the time shift information in the latent `time-encoded component' $\mathbf{D}_t$. To implement linear transformation along $\mathbf{D}_t$, we learn the coefficient (\ie, magnitude) on each direction of $\mathbf{D}_t$, noted as $ \mathbf{A} \in \mathbb{R}^{1 \times M} = \{a_{1}, a_{2}, ..., a_{M}\} = \et (\mathbf{dt}_q, \f (q))$, by the time encoder. This linear transformation is able to enforce time variance and to obtain different representations only along $\mathbf{D}_t$. The final time-blended representation $\fp (q, \mathbf{dt}_q)$ can be described as follows
\begin{equation}\label{ltn}
\small
\fp (q, \mathbf{dt}_q) = \f (q) + \sum^{M}_{i=1}a_{i} \cdot \mathbf{d_{i}} = \f (q) + \mathbf{A} \times \mathbf{D}^{T}_t .
\end{equation}
All proposed time parameterization variants are effective in learning video representations aware of temporal changes and can improve the target downstream tasks by capturing such motion variances. Associated analysis is presented in Sec.~Ablation Study, where we compare the three variants on their performance of downstream tasks. We find that the Linear Transformation with an orthogonal basis is the most effective and is beneficial as a generic methodology for learning time-aware spatio-temporal representations.  

\subsection{Self-supervised Contrastive Learning}\label{sec:contrastive}
\input{tables/ablation}

\input{tables/ablation2}


In this section, we omit the parameterized time of all samples in the notations to simplified formulations (\eg, $\fp(q, \mathbf{dt_q})$ is abbreviated as $\fp(q)$), and we provide details on the contrastive loss function. We apply general contrastive learning (see Fig.~\ref{fig:overview} (c)) to train our visual encoder $\f$ to encourage similarities between the time-blended positive representations, $\fp(q), \fp(k^{+}_1), ..., \fp(k^{+}_P)$, and discourage similarities between negative representations, $ \fp(k^{-}_1), ..., \fp(k^{-}_N)$. 
The InfoNCE~\cite{infonce} objective is defined as follows
\begin{equation}\label{loss}
\small
    \mathcal{L}_q = \sum_{p=1}^P \mathcal{L}_{NCE} = - \mathbb{E} \Big( \log  \frac{\sum_{p=1}^P e^ {\Sim\big ( \fp (q), \fp (k^{+}_p) \big )} }{ \sum_{n=1}^N e^{\Sim\big ( \fp (q), \fp(k^{-}_n) \big )}  } \Big),
\end{equation}
where $P$ represents the number of positive Keys, $N$ denotes the number of negative Keys, and the similarity can be computed as:
\begin{equation}\label{sim}
\small
    \Sim (x, y) =  \frac{\phi (x) \cdot \phi (y)}{\left\| \phi (x) \right\| \cdot \left\| \phi (y) \right\|} \cdot \frac{1}{Temp},
\end{equation}
where $Temp$ refers to the temperature hyper-parameter~\cite{non-para}, and $\phi$ is a learnable mapping function (\eg, an MLP projection head~\cite{study2021}) that can substantially improve the learned representations.

%% file: tables/ablation.tex
\begin{table}[t]
  \centering
     
     \scalebox{0.94}{
        \begin{tabular}{  l c c  c }
        
        \hline
        \multirow{1}*{\textbf{Transformation}}
        &\multirow{1}*{\text{Top-1 (\%)}} & 
           \multicolumn{1}{c}{\text{Mean (\%)}}\\
           \hline
           \hline
        \text{Base: w/o transformation} & 65.1& \text{49.7}\\
        \text{Variant 1: Linear w/o $\mathbf{D}_t$} & 66.0& \text{49.8}\\
        \text{Variant 2: Attention} &66.7  &\text{51.6} \\
        \hline
        \text{Variant 3: Linear w/ $\mathbf{D}_t$ }\\
        \text{~~w/o orthogonalization of $\mathbf{D}_t$} & 67.3& \text{53.1}\\
        \textbf{~~w/ orthogonalization of $\mathbf{D}_t$} &\textbf{67.8}  & \textbf{53.7}\\
        \hline
        \end{tabular}}
        \vspace{-0.2cm}
     \caption{Top-1 accuracy and Mean accuracy on Smarthome CS in comparing proposed Time parameterization variants. }
     \label{tab_app}
     \vspace{-0.cm}
\end{table}

\begin{table}[t]
   \centering

        \scalebox{0.94}{
        \setlength{\tabcolsep}{.8mm}{
        \begin{tabular}{  l c c c c }
        
        \hline
        \multirow{1}*{\textbf{Method}}
        &$P$
        &\multirow{1}*{\text{Top-1 (\%)}} & 
           \multicolumn{1}{c}{\text{Mean (\%)}} \\
           \hline
           \hline
        \text{MoCo~\cite{He_2020_CVPR}}& 2 & 61.5& \text{47.2}\\
        \text{$\rho$MoCo~\cite{study2021}}& 4 &65.1 & \text{49.7}\\
        \text{$\rho$BYOL~\cite{study2021}} &4 &\text{61.7}&  \text{42.4}\\
        \hline
        \text{LTN + MoCo}&2 &65.5 & \text{49.0}\\
        \text{LTN + $\rho$MoCo}&4 &\textbf{67.8}  & \textbf{53.7}\\
        \text{LTN + $\rho$BYOL}& 4&63.3 & \text{45.1}\\

        \hline
        \end{tabular}}}
        \vspace{-0.2cm}
        \caption{Top-1 accuracy and Mean per-class accuracy on Smarthome CS signifying the impact of LTN on \textit{different contrastive frameworks}. $P$: number of positive pairs.}
        \label{tab_ssl}
        \vspace{-0.3cm}
\end{table}

%% file: tables/ablation2.tex
\begin{table}[t]

  \centering
    
     \scalebox{0.94}{
        \begin{tabular}{  c c c c }
        
        \hline
        \multirow{1}*{\text{\#Layers}}
        &\multirow{1}*{\text{\#Dimensions}}
        &\multirow{1}*{\text{Top-1 (\%)}} & 
           \multicolumn{1}{c}{\text{Mean (\%)}}\\
           \hline
           \hline
        \text{None} & - & 65.1& \text{49.7}\\
        \hline
        \text{1} & 128 & 66.7& \text{50.5}\\
        \text{1} & 1024 & 67.3& \text{52.3}\\
        \text{2} & 1024 &67.1 & \text{52.8}\\
        \text{2} & 2048&\textbf{67.8}  & \textbf{53.7}\\
        \text{3} & 2048 &67.9 & \text{53.2}\\
        \hline
        \end{tabular}}
        \vspace{-0.2cm}
     \caption{Top-1 accuracy and Mean per-class accuracy on Smarthome CS \wrt \textit{Time Encoder}. }
     \label{tab_time}
     \vspace{-0.cm}
  
\end{table}

\begin{table}[t]
   \centering

        \scalebox{0.94}{
        \setlength{\tabcolsep}{3.mm}{
        \begin{tabular}{  c c c  }
        
        \hline
        \multirow{1}*{\textbf{Size of $\mathbf{D}_t$ ($M$)}}
        &\multirow{1}*{\text{Top-1 (\%)}} & 
           \multicolumn{1}{c}{\text{Mean (\%)}}\\
           \hline
           \hline
        \text{$M=16$} &65.2 & \text{51.6}\\
        \text{$M=64$} &\textbf{67.8}  & \textbf{53.7}\\
        \text{$M=128$} & 67.3 &\text{52.2} \\
        \text{$M=512$} & 67.6 &\text{52.1}\\
        \text{$M=1024$} &67.5 & \text{51.1}\\
        \text{$M=2000$} &66.9 & \text{50.5}\\

        \hline
        \end{tabular}}}
        \vspace{-0.2cm}
        \caption{Top-1 and Mean accuracy on Smarthome CS for study on number of directions in the orthogonal basis $\mathbf{D}_t$.}
        \label{tab_size}
        \vspace{-0.3cm}
\end{table}

%% file: sections/experiments.tex
\section{Experiments and Analysis}

\input{tables/sota-sh}

\input{tables/kinetics}

We conduct extensive experiments to evaluate LTN on four action classification datasets: \textbf{Toyota Smarthome}, \textbf{Kinetics-400}, \textbf{UCF101} and \textbf{HMDB51}.
Firstly, 
we provide experimental results on tested variants, we investigate exhaustive ablations and further analyze on Toyota Smarthome (fine-grained action classification dataset) to better understand the design choices of our proposed time parameterization approaches. Secondly, we compare LTN with the best setting to state-of-the-art methods on all evaluated benchmarks.
: Toyota Smarthome, UCF101 and HMDB51 without additional training data and with pre-training on Kinetics-400. 

\subsection{Ablation Study}\label{sec:ablation}

As activities of Toyota Smarthome (Smarthome) are with similar motion and high duration variance (\eg, `Leave', `Enter', `Clean dishes', `Clean up'), the temporal information is generally crucial for action classification. To understand the contribution of LTN for video representation learning, we conduct ablation experiments on Smarthome Cross-Subject~\cite{Das_2019_ICCV}, with \textit{linear evaluation} protocol (\ie, pre-training without action labels, then training the classifiers only with the action labels) using RGB videos without additional modalities or training data. For the proposed $\mathbf{D}_t$, unless otherwise stated, we set $M=64$ directions over the $dim=2048$ dimensions. We report Top-1 and Mean per-class accuracy.

\vspace{-.1cm}

\paragraph{LTN Variants.}
The key module of LTN is the Time Parameterization method with three effective variants 
To study the impact of each variant, we start from a baseline using MoCo~\cite{He_2020_CVPR} with multiple positive samples $P=4$ as~\cite{study2021} and we then incorporate the time parameterization variants. The results in Tab.~\ref{tab_app} indicate that leveraging time information is pertinent in improving the accuracy of fine-grained action classification. Specifically, in Variant 1, joint linear addition and visual representation related to time encoding without using $\mathbf{D}_t$ slightly boosts the Top-1 performance. We argue that the learned representation should code spatio-temporal data augmentation. If the entire representation is biased by time in the absence of $\mathbf{D}_t$, the static information that should be invariant is also shifted. This motivates us to use latent space decomposition to disentangle the `time-encoded component' $\mathbf{D}_t$ coded in the learned representation. Using $\mathbf{D}_t$ to parameterize time encoding can significantly improve the performance (+1.9\% by Variant 2 based on attention), especially by means of linear transformation (+4.0\% by Variant 3).

\vspace{-.1cm}
\paragraph{Impact of LTN for Different Contrastive Models.}
We compare two state-of-the-art momentum-based contrastive models~\cite{byol, He_2020_CVPR}, a pair of positive samples ($P$=2) and the improved versions~\cite{study2021} by leveraging multiple positive Keys ($P$=4) on the Smarthome dataset. Then, we incorporate the proposed LTN (Variant 3 with $M=64$) into all models. The results in Tab.~\ref{tab_ssl} demonstrate that LTN improves all three models and performs the best with $\rho$ MoCo~\cite{study2021} for our target downstream action classification task.

\vspace{-.1cm}
\paragraph{Design of Time Encoder.}
We explore how many directions are required in $\mathbf{D}_t$. We empirically test six different values for $M$ from 64 to 2000. Quantitative results in Tab.~\ref{tab_size} show
that when using 64 directions (out of all $dim$=2048 directions), the model achieves the best action classification results. Hence, we set $M=64$ for the other experiments. For the design of the proposed time encoder, we investigate the effect of different numbers of hidden layers and dimensions for the time encoder across five architectures. The results shown in Tab.~\ref{tab_time} suggest that 2-layer MLP with 2048 dimensions in the hidden layer is the most effective.

\subsection{Comparison with State-of-the-art}
\input{tables/perclasscom}

\input{tables/sota-k400}

\input{images/vis2}

We first compare our method on Smarthome. As we are the firsts to conduct the self-supervised action classification task on this dataset using only RGB data, we re-implement state-of-the-art models~\cite{SwAV, chen2020simclr, study2021, byol, He_2020_CVPR} and we compare the linear evaluation results without extra training data. We find that our proposed LTN, jointly with MoCo~\cite{He_2020_CVPR} achieves state-of-the-art performance, see Tab.~\ref{tab_sotash}. To further compare the results with skeleton-based methods~\cite{climent2021improved, das2020vpn} trained with additional stream~\cite{Yang_2021_WACV, unik}, we conduct a self-supervised pre-training on Kinetics-400 and we transfer the model on Smarthome by linear evaluation and fine-tuning, see Tab.~\ref{tab_sotash} bottom. In both settings, our model outperforms self-supervised state-of-the-art accuracy and many supervised approaches~\cite{climent2021improved, Das_2019_ICCV, das2020vpn, unseenview, Ryoo2020AssembleNetAM, 2sagcn2019cvpr}.

We then compare our method to state-of-the-art approaches by linear evaluation on the general video understanding benchmark, Kinetics-400. For fair comparison, we mainly focus on the methods using R3D-50 and $T=8$ sampled frames for training. The results are shown in Tab.~\ref{tab_lineark400} and demonstrate that, our LTN can improved upon previous methods~\cite{ study2021, He_2020_CVPR, Li_2021_ICCV, CVRL, tempo,  yao2021seco}. 

We also compare our LTN to state-of-the-art on HMDB51 and UCF101 (see Tab.~\ref{tab_sotak400}). For fair comparison, we mainly focus on the model trained with the R3D-50 backbone used in our work with training frames $T=8$. Using frozen features, our model outperforms all other works and even outperforms a number of works that adopt fine-tuning. For fine-tuning, the improvements are slight as the duration of these videos is small and the action is not as sensitive as Smarthome to time variance. However, we still outperform all previously single RGB-based models and our model performs competitively with current multi-modal methods~\cite{cotraining, Li_2021_ICCV, broaden_2021_ICCV} combining information from pre-extracted optical flow and audio. 

\subsection{Further Analysis}
\paragraph{Per-class Comparison with State-of-the-art.} We list the Smarthome classes that benefit the most and the least from LTN (see Tab.~\ref{tab_sh-bd}) compared to the state-of-the-art model (MoCo). We find that our method is able to effectively classify the fine-grained actions (\eg, `Cook.Usestove' +47.1\%, `Makecoffee.Boilwater' +31.8\%, `Laydown' +25.9\%, `Leave' +22.4\%) while being challenged in distinguishing some object-oriented activities (\eg, `Drink.Fromglass' -28.3\%, `Drink.Fromcan': -14.2\%). We believe that this is due to the fact that we focus on temporal modeling using time encoding, which only places emphasis on humans and ignores object information. To tackle this challenge and to further improve classification performance, future work will extend our method to latent spatial information~\cite{Sun_2021_ICCV} in order to capture the object information, while maintaining time awareness, which is still an open problem.

\vspace{-0.1cm}

\paragraph{Representation Analysis.} To demonstrate that the learned presentations are aware of temporal augmentations, we randomly select 2 videos (`Leave' and `Enter') that are correctly classified by our model and uniformly sample 20 segments for each video. Then, we visualize their time-aware (learned by the proposed LTN) and time-invariant (learned by MoCo) representations respectively with t-SNE (see Fig.~\ref{fig:vis2}). We find that, unlike the time-invariant representations of uniformly sampled segments learned by previous model~\cite{He_2020_CVPR} that are only regrouped together, the time-blended representations learned by our LTN are well aligned over the time order. Hence we conclude that LTN can learn the consistent amount of temporal changes with the video segments on their time-aware representations to benefit fine-grained motion-focused action classification.

%% file: tables/sota-sh.tex
\begin{table*}[t]
\centering

\begin{center}

\scalebox{0.92}{
\setlength{\tabcolsep}{1.2mm}{
\begin{tabular}{  l c c c c c c c }

\hline
\multirow{2}*{\textbf{Method}}
&\multirow{2}*{\text{Supervision}}
&\multirow{2}*{\text{Backbone}}
&\multirow{2}*{\text{Mod.}}
&\multirow{2}*{\text{Dataset}}
&\multirow{2}*{\text{Frozen}}
&\multicolumn{2}{c}{\textbf{Toyota Smarthome}}\\

& & & & & &\text{CS(\%)} &\text{CV2(\%)}\\
\hline
\hline
From scratch & Supervised & R3D-50& V& SH &  $\mathbf{\times}$ &50.2 &28.6  \\
\hline

SimCLR~\cite{chen2020simclr} & Self-sup. & R3D-50& V& SH &\textbf{\checkmark}&42.2 &26.3 \\
SwAV~\cite{SwAV} & Self-sup. & R3D-50& V& SH &\textbf{\checkmark}& 41.4 & 25.6\\
MoCo~\cite{He_2020_CVPR} & Self-sup. & R3D-50& V& SH &\textbf{\checkmark}& 47.2 &28.8 \\
$\rho$BYOL~\cite{study2021} & Self-sup. & R3D-50& V& SH & \textbf{\checkmark} & 42.4& 26.8 \\

\textbf{LTN (Ours)} & Self-sup. & R3D-50& V& SH& \textbf{\checkmark} &\textbf{53.7} &\textbf{30.1}\\
\textbf{LTN (Ours)} & Self-sup. & R3D-50& V& K400& \textbf{\checkmark} &\textbf{54.5} &\textbf{35.5}\\

\hline

\rowcolor{mygray}STA~\cite{Das_2019_ICCV}& Supervised & I3D+LSTM& V+P& K400 & $\mathbf{\times}$ &54.2 & 50.3\\
\rowcolor{mygray}AssembleNet++~\cite{Ryoo2020AssembleNetAM}& Supervised &R(2+1)D-50& V &K400 & $\mathbf{\times}$& \text{63.6} &- \\
\rowcolor{mygray}NPL~\cite{unseenview} & Supervised &R3D-50& V &K400 & $\mathbf{\times}$& - & \text{54.6}\\  
\rowcolor{mygray}ImprovedSTA~\cite{climent2021improved}& Supervised & I3D+LSTM& V+P& K400 & $\mathbf{\times}$ &63.7 & 53.6\\
\rowcolor{mygray} VPN~\cite{das2020vpn}& Supervised &I3D+AGCNs &V+P & K400& $\mathbf{\times}$& \text{60.8} &\text{53.5}\\

MoCo~\cite{He_2020_CVPR} & Self-sup. & R3D-50& V& K400 &$\mathbf{\times}$ &61.8  & 52.7\\
\textbf{LTN (Ours)} & Self-sup. & R3D-50& V& K400& $\mathbf{\times}$ &\textbf{65.9} &\textbf{54.6}\\

\hline

\end{tabular}}}
\end{center}
\vspace{-0.35cm}
\caption{Comparison of LTN to state-of-the-art methods on the Toyota Smarthome dataset (SH) with Cross-Subject (CS) and Cross-View2 (CV2) evaluation protocols. Mod: Modalities, V: RGB frames only, P: pre-extracted Pose data (skeleton keypoints coordinates), K400: the Kinetics-400 dataset. We classify methods \wrt supervision in the second column. }
\vspace{-0.cm}
\label{tab_sotash}
\end{table*}

%% file: tables/kinetics.tex
\begin{table}[t]
\centering
\begin{center}

\scalebox{0.92}{

\setlength{\tabcolsep}{.2mm}{
\begin{tabular}{  l c c c  c}

\hline
\multirow{1}*{\textbf{Method}}
&\multirow{1}*{\text{Backbone}}
&\multirow{1}*{\text{Mod.}}
&\multicolumn{1}{c}{\textbf{K400 (\%)}} \\
 \hline
\hline
VTHCL~\cite{tempo} &R3D-50 & V & 37.8\\
CVRL~\cite{CVRL}  & R3D-50 & V & 66.1 \\
SeCo~\cite{yao2021seco} & R3D-50 & V & 61.9\\
\text{MoCo~\cite{He_2020_CVPR}} & R3D-50 & V &\text{66.6} \\
\text{$\rho$BYOL~\cite{study2021}} & R3D-50 & V &\text{70.0} \\
\text{MCL~\cite{Li_2021_ICCV}} & R3D-50 & V+F &\text{66.6} \\
\hline
\textbf{LTN (Ours)} & R3D-50 & V &\textbf{71.3} \\

\hline

\end{tabular}}}
\end{center}
\vspace{-0.35cm}
\caption{Comparison with state-of-the-art methods on Kinetics-400 by \textit{Linear evaluation}. Mod: Modalities, V: RGB frames only, F: pre-extracted optical flow.}

\vspace{-0.3cm}
\label{tab_lineark400}
\end{table}

%% file: tables/perclasscom.tex
\begin{table}[t]
\centering

\begin{center}
\scalebox{0.92}{
\setlength{\tabcolsep}{6.8mm}{
\begin{tabular}{ c c }
\hline
\textbf{Activity}& \textbf{Gain from LTN (\%)} \\

\hline
\hline
\text{Cook.Usestove} & +47.08 \\
\text{Maketea.Boilwater} & +31.78 \\
\text{Laydown} & +25.88 \\
\text{Cutbread} & +25.42 \\
\text{Leave} & +22.43 \\
\text{Mean Accuracy} & +6.97 \\
\rowcolor{mygray}\text{Walk} & -5.07 \\
\rowcolor{mygray}\text{Usetablet} & -11.30 \\
\rowcolor{mygray}\text{Cook.Cleandishes} & -12.74 \\
\rowcolor{mygray}\text{Drink.Fromcan} & -14.24 \\
\rowcolor{mygray}\text{Drink.Fromglass} &-28.25 \\
\hline
\end{tabular}}}
\end{center}
\vspace{-0.35cm}
 
\caption{Activities that benefit the most and the least from LTN, and Mean per-class accuracy gain on Smarthome CS. }
\vspace{-0.3cm}
\label{tab_sh-bd}
\end{table}

%% file: tables/sota-k400.tex
\begin{table*}[t]
\centering

\begin{center}

\scalebox{0.92}{

\setlength{\tabcolsep}{.1mm}{
\begin{tabular}{  l c c c c c c | c c c c }

\hline
\multirow{1}*{\textbf{Method}}
&\multirow{1}*{\text{Backbone}}
&\multirow{1}*{\text{Mod.}}
&\multirow{1}*{\text{Data}}
&\multirow{1}*{\text{Frozen}}
&\textbf{UCF} (\%) &\textbf{HMDB} (\%)
&\multirow{1}*{\text{Data}}
&\multirow{1}*{\text{Frozen}}
&\textbf{UCF} (\%) &\textbf{HMDB} (\%)\\
\hline
\hline
OPN~\cite{opn} &VGG-M & V&-&\textbf{\checkmark}&-&-& UCF  &$\mathbf{\times}$ &59.6& 23.8\\
ClipOrder~\cite{cliporder} &R(2+1)D & V&-&\textbf{\checkmark}&-&-& UCF &$\mathbf{\times}$ &72.4& 30.9\\

CoCLR~\cite{cotraining} & S3D & V&UCF&\textbf{\checkmark}&70.2&39.1 & UCF &$\mathbf{\times}$& 81.4& 52.1 \\
\textbf{LTN (Ours)} & R3D-50& V& UCF& \textbf{\checkmark} &\textbf{71.8} &\textbf{40.3} &UCF & $\mathbf{\times}$&\textbf{81.6}& \textbf{52.8}\\
\hline
SpeedNet~\cite{speed} & S3D-G & V&-&\textbf{\checkmark}&-&- & K400 &$\mathbf{\times}$ & 81.1& 48.8\\
VTHCL~\cite{tempo}& R3D-50 & V&-&\textbf{\checkmark}&-&-& K400 & $\mathbf{\times}$ &82.1 & 49.2 \\
TaCo~\cite{taco} & R3D-50 & V& K400 &\textbf{\checkmark}& 59.6 & 26.7&  K400 &$\mathbf{\times}$ &85.1 & 51.6\\
MoCo~\cite{He_2020_CVPR} & R3D-50 & V&-&\textbf{\checkmark}&-&-& K400 &$\mathbf{\times}$& 92.8 & 67.5\\
CVRL~\cite{CVRL} & R3D-50 & V&-&\textbf{\checkmark}&-&-& K400 & $\mathbf{\times}$ &92.2 &66.7 \\ 
$\rho$BYOL~\cite{study2021} &R3D-50 & V&-&\textbf{\checkmark}&-&-& K400 & $\mathbf{\times}$ & 94.2& 72.1 \\
SeCo~\cite{yao2021seco} & R3D-50 & V& K400 &\textbf{\checkmark}& - & -& K400 & $\mathbf{\times}$& 88.3 & 55.6\\
CATE~\cite{Sun_2021_ICCV}& R3D-50 & V& K400 &\textbf{\checkmark}& 84.3 & 53.6 & K400 &$\mathbf{\times}$& 88.4 & 61.9 \\
CORP~\cite{corp_ICCV} & R3D-50 & V& K400 &\textbf{\checkmark}& 90.2& 58.7 & K400 &$\mathbf{\times}$& 93.5 &68.0 \\
FAME~\cite{ding2022motion} &I3D &V &K400 & \textbf{\checkmark}&-&- & K400 &$\mathbf{\times}$&  88.6& 61.1\\

\textbf{LTN (Ours)} & R3D-50 & V& K400& \textbf{\checkmark} &\textbf{90.6} &\textbf{58.9} & K400& $\mathbf{\times}$ &\textbf{94.5} &\textbf{72.3}\\

\rowcolor{mygray}CoCLR~\cite{cotraining} & S3D & V+F&K400&\textbf{\checkmark}&77.8&52.4& K400 &$\mathbf{\times}$& 90.6& 62.9 \\
\rowcolor{mygray}MCL~\cite{Li_2021_ICCV} & R(2+1)D-50 & V+F&-&\textbf{\checkmark}&-&-& K400 &$\mathbf{\times}$& 93.4& 69.1 \\
\rowcolor{mygray}BraVe~\cite{broaden_2021_ICCV} & TSM-50x2 & V+F+A&AudioS  &\textbf{\checkmark}&92.8&70.6 & AudioS &$\mathbf{\times}$& 96.5 &79.3  \\
\hline

\end{tabular}}}
\end{center}
\vspace{-0.35cm}
\caption{Comparison with state-of-the-art methods on UCF101 and HMDB51 with pre-training on Kinetics-400 (K400). Mod: Modalities, V: RGB frames only, F: pre-extracted optical flow, A: Audio.}

\vspace{-0.cm}
\label{tab_sotak400}
\end{table*}

%% file: images/vis2.tex
\begin{figure*}[t]
\begin{center}
\includegraphics[width=1\linewidth]{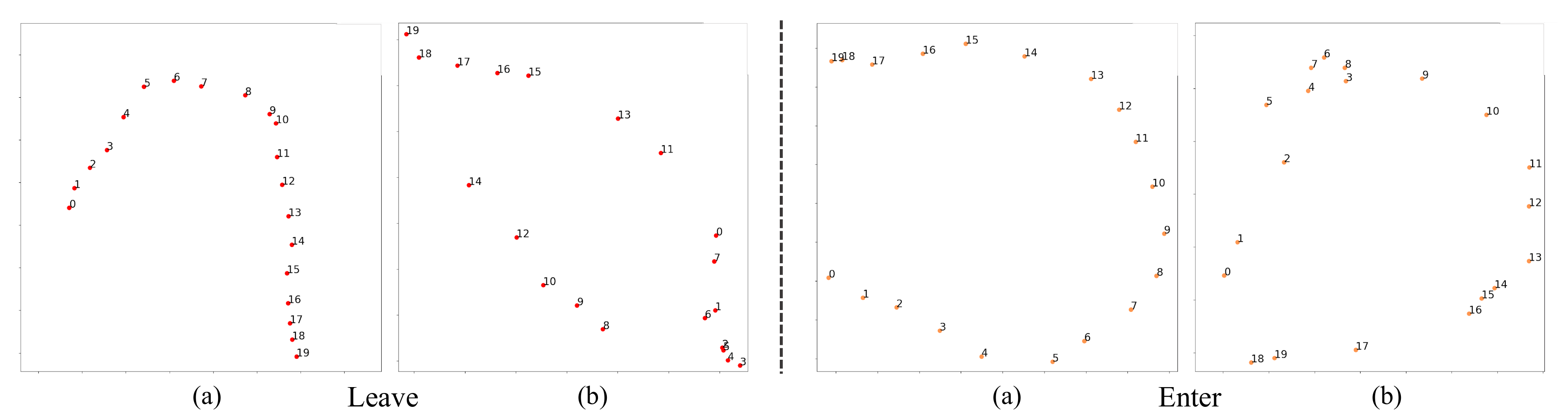}
\end{center}
\vspace{-0.5cm}
   \caption{\text{Impact of LTN} for video `Leave' and `Enter'. (a) Time-aware representations learned by LTN. (b) Their Time-invariant representations learned without LTN modules. The numbers indicate the time order of each uniformly sampled segment.}
\vspace{-.3cm}
\label{fig:vis2}
\end{figure*}

%% file: sections/conclusion.tex
\section{Conclusions}
In this work, we present LTN, a temporal parameterization approach that learns time-aware action representation.
We show that embedding time information of each video segment into the contrastive model by time navigation through a time encoder and an orthogonal basis can significantly improve the representation capability for videos. Experimental analysis confirms that a visual encoder extracting such representation can boost downstream action recognition.
Future work will extend our time parameterization approach to spatial dimension, in order to better capture the object information that may also be crucial for fine-grained action recognition.

\section*{Acknowledgements} This work was supported by Toyota Motor Europe (TME) and the French government, through the 3IA Cote d’Azur Investments In the Future project managed by the National Research Agency (ANR) with the reference number ANR-19-P3IA-0002. This work was granted access to the HPC resources of IDRIS under the allocation AD011011627R1.